\documentclass{article} 
\usepackage{nips13submit_e,times}
\usepackage{hyperref}
\usepackage{url}
\usepackage{graphicx,amssymb,amsmath}

\title{End-to-end Phoneme Sequence Recognition using Convolutional Neural Networks}

\author{Dimitri Palaz$^{1,2}$, Ronan Collobert$^1$, Mathew Magimai.-Doss$^1$ \\
$^1$Idiap Research Institute, Martigny, Switzerland \\ $^2$Ecole Polytechnique F\'ed\'erale de Lausanne (EPFL), Lausanne, Switzerland\\
{\small \tt \{dimitri.palaz, ronan.collobert, mathew\}@idiap.ch}
}

\nipsfinalcopy 

\begin{document}

\maketitle

\begin{abstract}
Most phoneme recognition state-of-the-art systems rely on a classical neural network classifiers, fed with highly tuned features, such as MFCC or PLP features. Recent advances in ``deep learning'' approaches questioned such systems, but while some attempts were made with simpler features such as spectrograms, state-of-the-art systems still rely on MFCCs. This might be viewed as a kind of failure from deep learning approaches, which are often claimed to have the ability to train with raw signals, alleviating the need of hand-crafted features. In this paper, we investigate a convolutional neural network approach for raw speech signals. While convolutional architectures got tremendous success in computer vision or text processing, they seem to have been let down in the past recent years in the speech processing field. We show that it is possible to learn an end-to-end phoneme sequence classifier system directly from raw signal, with similar performance on the TIMIT and WSJ datasets than existing systems based on MFCC, questioning the need of complex hand-crafted features on large datasets. 
\end{abstract}

\section{Introduction}

Most of the state-of-the-art systems in phoneme recognition system tend to follow
the same approach: the task is divided in several sub-tasks, which are
optimized in an independent manner. In a first step, the data is
transformed into features, usually composed of a dimensionality reduction
phase and an information selection phase, based on the task-specific
knowledge of the phenomena.  These two steps have been carefully
hand-crafted, leading to state-of-the-art features such as MFCCs
or PLPs. In a second step, the likelihood of sub-sequence units is
estimated using generative or discriminative models, making several assumptions, for example
units form a Markov chain. In a final step, dynamic programming techniques
are used to recognize the sequence under constraints.

Recent advances in machine learning have made possible systems that can be
trained in an end-to-end manner, i.e. systems where every step is
\textit{learned} simultaneously, taking into account all the other steps
and the final task of the whole system. It is usually referred as
\textit{deep learning}, mainly because such architectures are usually
composed of many layers (supposed to provide an increasing level of
abstraction), compared to classical ``shallow'' systems. As opposed to
``divide and conquer'' approaches presented previously (where each step in
independently optimized) deep learning approaches are often claimed to have
the potential to lead to more optimal systems, and to have the advantage to
alleviate the need of find the right features for a given task of
interest. While there is a good success record of such approaches in the
computer vision \cite{lecun_gradient-based_1998} or text processing fields
\cite{collobert_natural_2011}, most speech systems still rely on complex
features such as MFCCs, including recent advanced deep learning
approaches~\cite{mohamed_acoustic_2012}. This contradicts the claim that
these techniques have end-to-end learning potential.


In this paper, we propose an end-to-end phoneme sequence recognition system, taking directly raw speech signal as inputs and which outputs a phoneme sequence. The system is composed of two parts: Convolutional Neural Networks~\cite{lecun_generalization_1989} (CNNs) perform the feature learning and classification stages, and a simple version of Conditional Random Field (CRFs) is used for the decoding stage, trained in a end-to-end manner. We show that such system can in fact lead to competitive
phoneme recognition systems, even when trained on raw signals. In the framework of hybrid HMM/ANN system, we compare the proposed approach 
with the conventional approach of extracting spectral-based acoustic feature extraction and then modeling them by ANN. Experimental studies conducted on TIMIT and WSJ corpus show that the proposed approach can yield a phoneme recognition system that is similar to or better than the system based on conventional approach.

The remainder of the paper is organized as follows. Section~\ref{sec-related-work} presents a brief survey of related literature. Section~\ref{sec-baseline} presents the classical HMM/ANN system. Section~\ref{sec-system} presents the architecture of the proposed system. Section~\ref{sec-setup} presents the experimental setup and Section~\ref{sec-results} presents the results and the discussion. Section~\ref{sec-conclusion} concludes the paper.



\section{Related Work}
\label{sec-related-work}
Deep learning architectures have been successfully applied to a wide range of application: characters recognition \cite{lecun_gradient-based_1998}, object 
recognition \cite{lecun_learning_2004}, natural language processing \cite{collobert_unified_2008} or image classification \cite{krizhevsky_imagenet_2012}

In speech, one of the first phoneme recognition system based on neural network was the Time Delay Neural Network \cite{waibel_phoneme_1989}. 
It was extended to isolated word recognition \cite{bottou_experiments_1989}. 
At the same time, the 
hybrid HMM/ANN architecture approach \cite{bourlard_links_1990,bengio_connectionist_1993} was developed, leading to more scalable systems. 

Recently, the deep belief network \cite{hinton_fast_2006} approach has been found to yield good performance in phone recognition \cite{mohamed_deep_2009}. It was also extend to context-dependent phonemes in \cite{seide_conversational_2011}. However, these systems used complex hand-crafted features, such as MFCC. Later, there has been growing interests in using short-term spectrum as features. These ``intermediate'' representations (standing between raw signal and ``classical'' features such as cepstral-based features) have been successfully used in speech recognition applications, for example in unsupervised feature learning~\cite{lee_unsupervised_2009}, acoustic modeling~\cite{mohamed_acoustic_2012} and large vocabulary speech recognition~\cite{dahl_context-dependent_2012,hinton_deep_2012}. Convolutional neural networks also yield state-of-the-art results in phoneme recognition~\cite{abdel-hamid_applying_2012}. Although these systems are able to learn efficient representations, these features are still used as input for hybrid systems. 

The work presented in~\cite{jaitly_learning_2011} successfully investigates features learning from raw speech for phoneme recognition. Here also, these features are used as input for a different system. Finally, convolutional neural networks have shown the capability of learning features from raw speech and estimate phoneme conditional probabilities in a single system~\cite{Palaz_INTERSPEECH_2013}.

\section{Phoneme sequence recognition using hybrid HMM/ANN system}
\label{sec-baseline}
\begin{figure}[h]
\begin{center}
\includegraphics{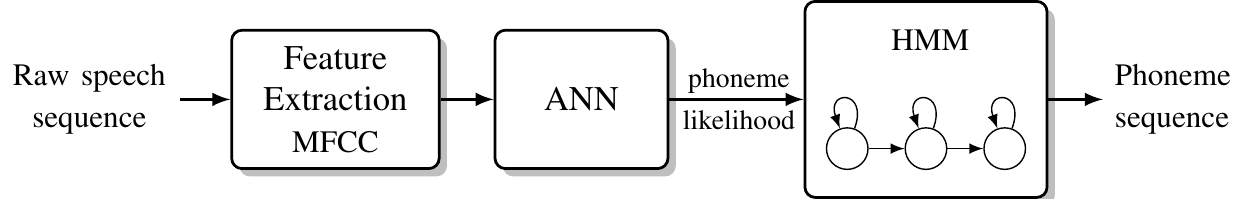}
\caption{Hybrid ANN/HMM phoneme recognition}
\label{fig:hybrid}
\end{center}
\end{figure}

The hybrid HMM/ANN system is one of the most common system for phoneme recognition, presented in Figure~\ref{fig:hybrid}. It is composed of three parts: features extraction, classification and decoding. In the first step, features are extracted from the signal, by transformation and filtering. The most common ones are Mel Frequency Cepstrum Coefficients (MFCC)~\cite{davis_comparison_1980} and Perceptual Linear Prediction (PLP)~\cite{hermansky1990perceptual}. Usually, the first and second derivative of these representations are computed over several surrounding frames and used as additional features. They are then given as input to a Artificial Neural Network (ANN), along with a 4 frames left and 4 frames right context. The network is usually a  feed-forward MLP composed of a hidden layer and a output layer, which computes the conditional probabilities for each class. These probabilities are then used as emission probabilities in a Hidden Markov Model (HMM) framework, which decodes the sequence. 

\section{Proposed system}
\label{sec-system}
\begin{figure}[h]
\begin{center}
\includegraphics{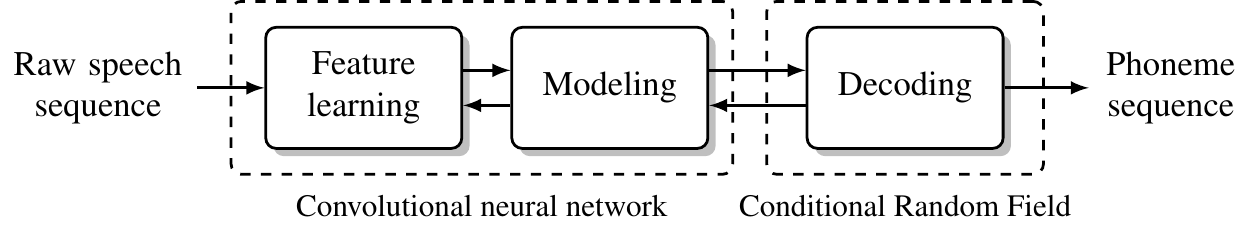}
\caption{End-to-end phoneme recognition system}
\label{fig:overview}
\end{center}
\end{figure}

The proposed system is composed of three stages: the feature learning stage, the modeling stage, which performs the classification, and the decoding of the sequence, as presented in Figure~\ref{fig:overview}. CNNs are used for the first two stages. Their \emph{convolutional} aspect make them particularly suitable for
 handling temporal signals such as raw speech. Moreover, the addition of max-pooling layers, classical layers in deep architecture, brings robustness against temporal distortion to the system and helps control the network capacity. For the third stage, a decoder based on CRFs is proposed, which will learn the transition between the different classes. The back-propagation algorithm is used for training the whole system in an end-to-end manner.

\subsection{Convolutional Neural Network}
\begin{figure}[h]
\begin{center}
\includegraphics[width=\textwidth]{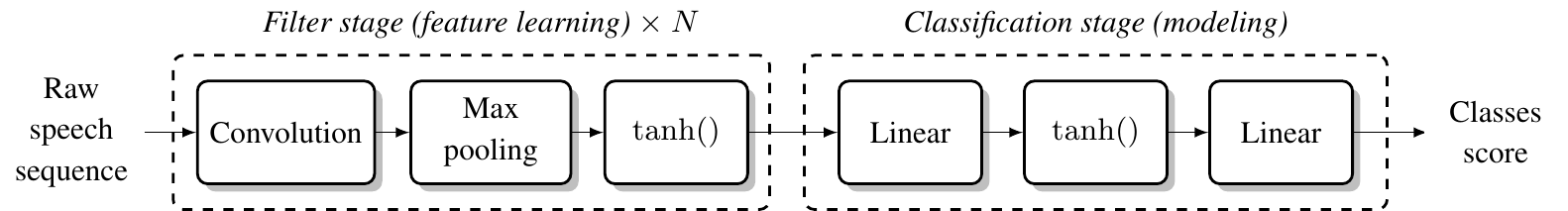}
\caption{{\it Convolutional Neural Network for one frame classification. Several stages of convolution/pooling/tanh might be
considered. Our network included 3 stages.}}
\label{fig:net}
\end{center}
\end{figure}

The network is given a sequence of raw input signal, split into frames, and outputs a score for each classes, for each frame. These type of network architectures are
composed of several filter extraction stages, followed by a classification
stage. A filter extraction stage involves a convolutional layer, followed
by a temporal pooling layer and an non-linearity ($\tanh()$). Our optimal
architecture included 3 stages of filter extraction (see Figure~\ref{fig:net}).
Signal coming out of these stages are fed to a classification stage,
which in our case is two linear layers with a large number of hidden units.

\subsubsection{Convolutional layer}
\begin{figure}[h]
\begin{minipage}{0.46\textwidth}
\begin{center}
\includegraphics[width=0.9\textwidth]{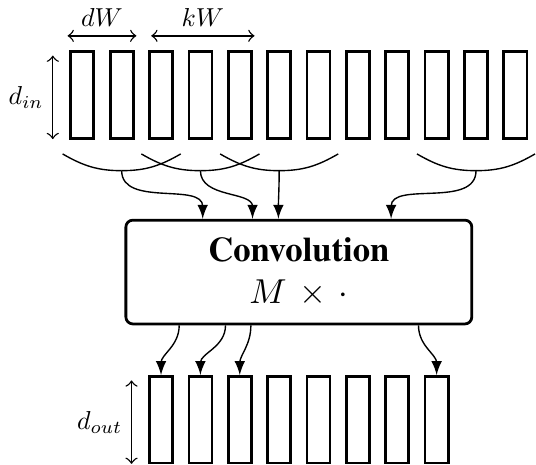}
\end{center}
\caption{\it Illustration of a convolutional layer. $d_{in}$ and $d_{out}$ are the dimension of the input and output frames. $kW$ is the kernel width (here $kW=3$) and $dW$ is the shift between two linear applications (here, $dW=2$).}  
\label{fig:cnn}
\end{minipage}
\hfill
\begin{minipage}{0.46\textwidth}
\begin{center}
\includegraphics[width=0.9\textwidth]{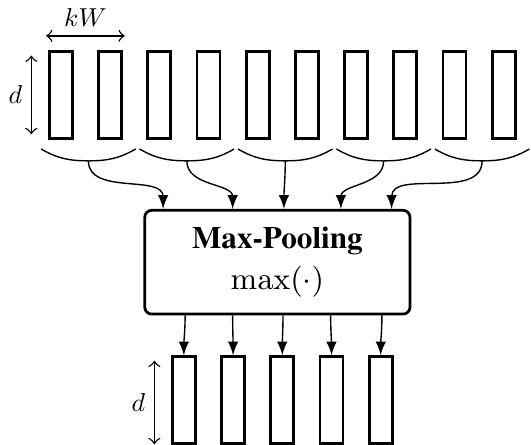}
\end{center}
\caption{{\it Illustration of max-pooling layer. $kW$ is the number of frame taken for each $\max$ operation and $d$ represents the dimension of input/output frames (which are equal).}}  
\label{fig:mp}
\end{minipage}
\end{figure}

While ``classical'' linear layers in standard MLPs accept a fixed-size input vector,
a convolution layer is assumed to be fed with a sequence of $T$ vectors/frames:
$X = \{x^1 \;\; x^2 \;\; \ldots \;\; x^T\}$.  A convolutional layer applies the same linear
transformation over each successive (or interspaced by $dW$ frames) windows of $kW$ frames.
E.g, the transformation at frame $t$ is formally written as:
\begin{equation}
M \left( 
\begin{array}{c}  
x^{t-(kW-1)/2} \\ \vdots \\ x^{t+(kW-1)/2}
\end{array} 
\right)\,,
\end{equation}
where $M$ is a $d_{out}\times d_{in}$ matrix of parameters.
In other words, $d_{out}$ filters (rows of the matrix M) are applied
to the input sequence.
An illustration is provided in Figure \ref{fig:cnn}.

\subsubsection{Max-pooling layer}
These kind of layers perform local temporal $\max$ operations over an input
sequence, as shown in Figure \ref{fig:mp}. More formally, the transformation
at frame $t$ is written as:
\begin{equation}
\underset{t-(kW-1)/2 \leq s \leq t+(kW-1)/2}{\max} \ x_{s}^i  \quad\quad \forall i
\end{equation}
These layers increase the robustness of the network to slight temporal
distortions in the input.


\subsection{Decoding}
\label{sec-crf}
\begin{figure}[t]
\begin{center}
\includegraphics{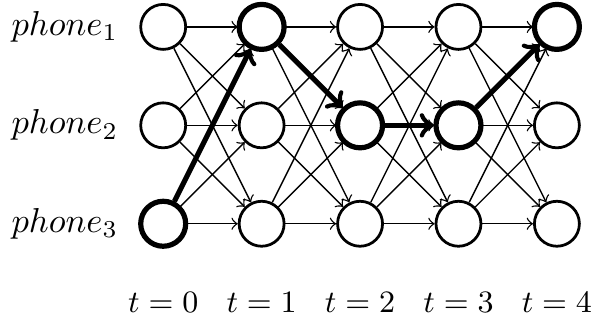}
\end{center}
\caption{{\it Illustration of the CRF graph.}}  
\label{fig:crf}
\end{figure}

We consider a simple version of CRFs, where we define a graph with
nodes for each frame in the input sequence, and each label. This CRF allows to discriminatively train a simple duration model over our network
output scores. Transition scores are assigned to edges between phonemes,
and network output scores are assigned to nodes. Given an input data
sequence $[x]_1^T$ and a label path on the graph $[i]_1^T$, a score for the path can be
defined:
\begin{equation}
s([x]_1^T,[i]_1^T)=\displaystyle \sum_{t=1}^T \left(f_{i_t}(x_t) + A_{i_t,i_{t-1}}\right)
\label{eq:crfscore}
\end{equation}
where $A$ is a matrix describing transitions between labels and
$f_{i_t}(x_t)$ the network score of input $x$ for class $i$ at time $t$, as illustrated in Figure~\ref{fig:crf}. Note that this approach is a particular case of the forward training of Graph Transformer Network~\cite{bottou_global_1997}.

At inference time, given a input sequence $[x]_1^T$, the best label path can be found by minimizing~\eqref{eq:crfscore}. The Viterbi algorithm is used to find
\begin{equation}
\underset{[j]_1^T}{\operatorname{argmax}}(s([x]_1^T,[j]_1^T,\theta))
\end{equation}

\subsection{Network training}
The network parameters $\theta$ are learned by maximizing the log-likelihood $L$, given by:
\begin{equation}
L(M_1,...,M_L,\theta)=\displaystyle \sum_{n=1}^N\log(p([i]_1^T|[x]_1^T,\theta)) 
\label{eq:like}
\end{equation}
for each input speech sequence $[x]_1^T$ and label sequence $[i]_1^t$, over the whole training set, with respect to the parameters of each layer $M_l$. Defining the $\operatorname{logadd(\cdot)}$ operation as: 
\begin{equation}
\label{eq:logadd}
\underset{i}{\operatorname{logadd}}(z_i)=\log(\displaystyle\sum_i e^{z_i})
\end{equation}
the likelihood $L$ can be expressed as:
\begin{equation}
L=\log(p([i]_1^T|[x]_1^T))=s([x]_1^T,[i]_1^T,\theta) - \underset{j}{\operatorname{logadd}}(s([x]_1^T,[j]_1^T,\theta))
\label{eq:likeframe}
\end{equation}
where $s([x]_1^T,[i]_1^T,\theta)$ is the path score as defined in~\eqref{eq:crfscore}. The number of term in the logadd operation grows exponentially with the length of the input sequence. Using a recursive algorithm, the logadd can be computed in linear time as presented in~\cite{collobert_natural_2011}.
The likelihood is then maximized using the stochastic gradient ascent algorithm \cite{bottou_stochastic_1991}.



\section{Experimental Setup}
\label{sec-setup}
In this section we present the setup used for the experiments, the databases and the hyper-parameters of the networks.

\subsection{Databases}
\subsubsection{TIMIT Corpus}
The TIMIT acoustic-phonetic corpus consists of 3,696 training utterances (sampled at 16kHz) from 462 speakers, excluding the SA sentences. The cross-validation set consists of 400 utterances from 50 speakers. The core test set was used to report the results. It contains 192 utterances from 24 speakers, excluding the validation set. Three different sets of phoneme were used: the first one is composed of 39 phonemes, as presented in \cite{lee_speaker-independent_1989}. The second one has 117 classes (3 states of each of the 39 phonemes). The last one uses 183 classes labels (3 states for each one of the original 61 phonemes). In the last two cases, after decoding, the classes were mapped to the 39 phonemes set for evaluation.

\subsubsection{Wall Street Journal Corpus}
The SI-284 set of the corpus~\cite{woodland_large_1994} is selected for the experiments. The set contains 36416 sequences, representing around 80 hours of speech. Ten percent of the set was taken as validation set. The ``Hub~2~2.5k'' set was selected as test set. It contains 215 sequences from 10 speakers. The phoneme sequences were extracted from the transcript and segmented using trained GMM. The CMU phoneme set, containing 40 classes, was used.

\subsection{Baseline system}
The baseline system is a standard HMM/ANN system~\cite{morgan_continuous_1995}. For the features, MFCC were used. They were computed (with HTK \cite{htk}) using a 25 ms Hamming window on the
speech signal, with a shift of 10 ms. The signal is represented using
13th-order coefficients along with their first and second derivatives,
computed on a 9 frames context. The classifier is a two-layer MLP. The decoding of the sequence was performed by a standard HMM
decoder, with constrained duration of 3 states, and considering all phoneme
equally probable.

\subsection{Network hyper-parameters}
\label{sec-tuning}
%

The hyper-parameters of the network are: the input window size, corresponding to the context taken along with each example, the number for sample for each example, the kernel width $kW$ and shift $dW$ of the convolutions, the number of filters $d_{out}$, the width of the hidden layer and the pooling width. They were tuned by early-stopping on the cross-validation set. Ranges which were considered for the grid search are reported in Table \ref{tab:hp}. It is worth
mentioning that for a given input window size over the raw signal, the size
of the output of the filter extraction stage will strongly depend on the
number of max-pooling layers, each of them dividing the output size of the
filter stage by the chosen pooling kernel width. As a result, adding
pooling layers \emph{reduces} the input size of the classification stage,
which in returns reduces the number of parameters of the network (as most
parameters do lie in the classification stage).

The best performance for TIMIT corpus on the cross-validation set for the first phoneme set was found with: $5$ ms duration for each example, $100$ ms of context, $10$, $3$ and $9$ frames kernel width, $10$, $1$ and $1$ frames shift, $100$ filters, $500$ hidden units and $3$ pooling width. For the second set: $10$ ms duration for each example, $100$ ms of context, $10$, $5$ and $7$ frames kernel width, $10$, $1$ and $1$ frames shift, $100$ filters, $500$ hidden units and $4$ pooling width. For the third set: $7.5$ ms duration for each example, $150$ ms of context, $10$, $7$ and $7$ frames kernel width, $10$, $1$ and $1$ frames shift, $100$ filters, $500$ hidden units and $2$ pooling width.
For the WSJ corpus: $10$ ms duration for each example, $680$ ms of context, $10$, $7$ and $9$ frames kernel width, $10$, $1$ and $1$ frames shift, $100$ filters, $1000$ hidden units and $2$ pooling width.
For the baseline, early stopping on the cross-validation set was also used to determine the optimal number of nodes ($500$ and $1000$ nodes were found).
The experiments were implemented using the \emph{torch7} toolbox \cite{collobert_torch7_2011}. 
 
\begin{table}[t]
\caption{Network hyper-parameters}
\label{tab:hp}
\begin{center}
\begin{tabular}{ll}
\multicolumn{1}{c}{\bf Parameters}  &\multicolumn{1}{c}{\bf Range}
\\ \hline \\
Input window size (ms) & 100-700 \\
Example duration (ms) & 5-15 \\
Kernel width ($kW$) & 1-9\\
Number of filters per kernel ($d_{out}$) & 10-90 \\
Number of hidden units in the class. stage & 100-1500 \\
\end{tabular}
\end{center}
\end{table}

\section{Results and discussion}
\label{sec-results}
The results are given in term of phoneme recognition accuracy, which is the Levenstein distance between the reference and the inference phoneme sequence. They are presented in Table~\ref{tab:timit} for the TIMIT corpus for the three phoneme sets, along with the number of parameters of the network. For comparison purposes, the accuracy using larger phoneme set is computed by mapping the classes to the first phoneme set. The proposed system yields better performance than the baseline on the TIMIT database. Using a larger number of classes and three states by phoneme also improves the performance. The results for the larger database, the WSJ corpus, are presented in Table~\ref{tab:wsj}. The proposed system also yields similar performances to the baseline, showing its scalability capacities.

The key difference between systems is that, for the proposed system, almost no prior knowledge on the data was used (except that the data is a temporal signal) and still achieve similar performances. This suggest that the deep network can learn relevant features, thus questioning the use of complex hand-crafted features. Moreover, by learning the transition between phoneme, the CRF is able to learn a duration model directly on training data, without any external constraints.

The end-to-end aspect of the proposed system makes it interesting for a stand-alone implementation of a phoneme recognizer, as the system takes sequences of raw speech as input and outputs phoneme sequences. Moreover, the speed of the system at inference makes it suitable for real-time phoneme recognition. A demo will be shown at the time of the conference.

\begin{table}[t]
\caption{Phoneme recognition accuracy on the core test set of TIMIT corpus.}
\label{tab:timit}
\begin{center}
\begin{tabular}{llll}
\multicolumn{1}{c}{\bf System}  &\multicolumn{1}{c}{\bf \#Classes}  &\multicolumn{1}{c}{\bf \#Param.} &\multicolumn{1}{c}{\bf Test acc.}\\
\hline \\
Baseline & 39 & 196'040 & 66.65 \%\\
CNN+CRF & 39 & 873'340 & 65.81 \%\\
CNN+CRF & 117 & 986'680 & 67.84 \%\\
CNN+CRF & 183 & 803'363 & 70.08 \%\\
\hline
\end{tabular}
\end{center}
\end{table} 

\begin{table}[t]
\caption{Phoneme recognition accuracy on the `Hub 2 2.5k'' test set of WSJ corpus.}
\label{tab:wsj}
\begin{center}
\begin{tabular}{llll}
\multicolumn{1}{c}{\bf System}  &\multicolumn{1}{c}{\bf \#Classes} &\multicolumn{1}{c}{\bf \#Param.} &\multicolumn{1}{c}{\bf Test acc.}\\
\hline \\
Baseline & 39 & 1'786'440 & 72.39 \%\\
CNN+CRF & 39 & 6'573'440 & 72.88 \% \\
\hline
\end{tabular}
\end{center}
\end{table} 

\section{Conclusion}
\label{sec-conclusion}
In this paper, we proposed an end-to-end phoneme recognition system, which is able to learn the feature by taking raw speech data as input and yield similar performances as baseline systems. As future work, we plan to improve the current system by investigating deeper architectures or constrained CRF. We will also extend it to context-dependent phonemes, therefore having more classes, which might lead to better performances, as Table~\ref{tab:timit} suggests. From there, we will focus on developing more specific applications, such as Spoken Term Detection.

\subsubsection*{Acknowledgments}
This work was partly supported by the HASLER foundation through the grant ``Universal Spoken Term Detection with Deep Learning'' (DeepSTD) and by the Swiss NSF through the Swiss National Center of Competence in Research (NCCR) on Interactive Multimodal Information Management (www.im2.ch).
\small
\bibliography{biblio}

\begin{thebibliography}{10}
\providecommand{\url}[1]{#1}
\csname url@samestyle\endcsname
\providecommand{\newblock}{\relax}
\providecommand{\bibinfo}[2]{#2}
\providecommand{\BIBentrySTDinterwordspacing}{\spaceskip=0pt\relax}
\providecommand{\BIBentryALTinterwordstretchfactor}{4}
\providecommand{\BIBentryALTinterwordspacing}{\spaceskip=\fontdimen2\font plus
\BIBentryALTinterwordstretchfactor\fontdimen3\font minus
  \fontdimen4\font\relax}
\providecommand{\BIBforeignlanguage}[2]{{%
\expandafter\ifx\csname l@#1\endcsname\relax
\typeout{** WARNING: IEEEtran.bst: No hyphenation pattern has been}%
\typeout{** loaded for the language `#1'. Using the pattern for}%
\typeout{** the default language instead.}%
\else
\language=\csname l@#1\endcsname
\fi
#2}}
\providecommand{\BIBdecl}{\relax}
\BIBdecl

\bibitem{lecun_gradient-based_1998}
Y.~{LeCun}, L.~Bottou, Y.~Bengio, and P.~Haffner, ``Gradient-based learning
  applied to document recognition,'' \emph{Proceedings of the {IEEE}}, vol.~86,
  no.~11, pp. 2278--2324, 1998.

\bibitem{collobert_natural_2011}
R.~Collobert, J.~Weston, L.~Bottou, M.~Karlen, K.~Kavukcuoglu, and P.~Kuksa,
  ``Natural language processing (almost) from scratch,'' \emph{The Journal of
  Machine Learning Research}, vol.~12, pp. 2493--2537, 2011.

\bibitem{mohamed_acoustic_2012}
A.~Mohamed, G.~Dahl, and G.~Hinton, ``Acoustic modeling using deep belief
  networks,'' \emph{Audio, Speech, and Language Processing, IEEE Transactions
  on}, vol.~20, no.~1, pp. 14 --22, jan. 2012.

\bibitem{lecun_generalization_1989}
Y.~{LeCun}, ``Generalization and network design strategies,'' in
  \emph{Connectionism in Perspective}, R.~Pfeifer, Z.~Schreter, F.~Fogelman,
  and L.~Steels, Eds.\hskip 1em plus 0.5em minus 0.4em\relax Zurich,
  Switzerland: Elsevier, 1989.

\bibitem{lecun_learning_2004}
Y.~{LeCun}, F.~J. Huang, and L.~Bottou, ``Learning methods for generic object
  recognition with invariance to pose and lighting,'' in \emph{Proceedings of
  the {IEEE} Computer Society Conference on Computer Vision and Pattern
  Recognition}, vol.~2, 2004, pp. II--97.

\bibitem{collobert_unified_2008}
R.~Collobert and J.~Weston, ``A unified architecture for natural language
  processing: deep neural networks with multitask learning,'' in
  \emph{Proceedings of the 25th international conference on Machine learning},
  2008, pp. 160--167.

\bibitem{krizhevsky_imagenet_2012}
A.~Krizhevsky, I.~Sutskever, and G.~Hinton, ``Imagenet classification with deep
  convolutional neural networks,'' in \emph{Advances in Neural Information
  Processing Systems 25}, 2012, pp. 1106--1114.

\bibitem{waibel_phoneme_1989}
A.~Waibel, T.~Hanazawa, G.~Hinton, K.~Shikano, and K.~Lang, ``Phoneme
  recognition using time-delay neural networks,'' \emph{Acoustics, Speech and
  Signal Processing, IEEE Transactions on}, vol.~37, no.~3, pp. 328 --339, mar
  1989.

\bibitem{bottou_experiments_1989}
L.~Bottou, F.~Fogelman~Souli\'e, P.~Blanchet, and J.~S. Lienard, ``Experiments
  with time delay networks and dynamic time warping for speaker independent
  isolated digit recognition,'' in \emph{Proceedings of {EuroSpeech} 89},
  vol.~2, Paris, France, 1989, pp. 537--540.

\bibitem{bourlard_links_1990}
H.~Bourlard and C.~Wellekens, ``Links between markov models and multilayer
  perceptrons,'' \emph{{IEEE} Transactions on Pattern Analysis and Machine
  Intelligence}, vol.~12, no.~12, pp. 1167 --1178, Dec. 1990.

\bibitem{bengio_connectionist_1993}
Y.~Bengio, ``A connectionist approach to speech recognition,''
  \emph{International Journal on Pattern Recognition and Artificial
  Intelligence}, vol.~7, no.~4, pp. 647--668, 1993.

\bibitem{hinton_fast_2006}
G.~E. Hinton, S.~Osindero, and Y.~W. Teh, ``A fast learning algorithm for deep
  belief nets,'' \emph{Neural computation}, vol.~18, no.~7, pp. 1527--1554,
  2006.

\bibitem{mohamed_deep_2009}
A.~Mohamed, G.~Dahl, and G.~Hinton, ``Deep belief networks for phone
  recognition,'' in \emph{{NIPS} Workshop on Deep Learning for Speech
  Recognition and Related Applications}, 2009.

\bibitem{seide_conversational_2011}
F.~Seide, G.~Li, and D.~Yu, ``Conversational speech transcription using
  context-dependent deep neural networks,'' in \emph{Proc. Interspeech}, 2011,
  pp. 437--440.

\bibitem{lee_unsupervised_2009}
H.~Lee, P.~Pham, Y.~Largman, and A.~Y. Ng, ``Unsupervised feature learning for
  audio classification using convolutional deep belief networks,'' in
  \emph{Advances in Neural Information Processing Systems 22}, 2009, pp.
  1096--1104.

\bibitem{dahl_context-dependent_2012}
G.~E. Dahl, D.~Yu, L.~Deng, and A.~Acero, ``Context-dependent pre-trained deep
  neural networks for large-vocabulary speech recognition,'' \emph{Audio,
  Speech, and Language Processing, {IEEE} Transactions on}, vol.~20, no.~1, p.
  30–42, 2012.

\bibitem{hinton_deep_2012}
G.~Hinton, L.~Deng, D.~Yu, G.~E. Dahl, A.-r. Mohamed, N.~Jaitly, A.~Senior,
  V.~Vanhoucke, P.~Nguyen, and T.~N. Sainath, ``Deep neural networks for
  acoustic modeling in speech recognition: the shared views of four research
  groups,'' \emph{Signal Processing Magazine, {IEEE}}, vol.~29, no.~6, p.
  82–97, 2012.

\bibitem{abdel-hamid_applying_2012}
O.~Abdel-Hamid, A.-r. Mohamed, H.~Jiang, and G.~Penn, ``Applying convolutional
  neural networks concepts to hybrid {NN-HMM} model for speech recognition,''
  in \emph{Proc. of ICASSP}, 2012, pp. 4277--4280.

\bibitem{jaitly_learning_2011}
N.~Jaitly and G.~Hinton, ``Learning a better representation of speech
  soundwaves using restricted boltzmann machines,'' in \emph{Proc. of ICASSP},
  2011, pp. 5884--5887.

\bibitem{Palaz_INTERSPEECH_2013}
D.~Palaz, R.~Collobert, and M.~Magimai.-Doss, ``Estimating phoneme class
  conditional probabilities from raw speech signal using convolutional neural
  networks,'' in \emph{Proceedings of Interspeech}, Aug. 2013.

\bibitem{davis_comparison_1980}
S.~Davis and P.~Mermelstein, ``Comparison of parametric representations for
  monosyllabic word recognition in continuously spoken sentences,''
  \emph{Acoustics, Speech and Signal Processing, {IEEE} Transactions on},
  vol.~28, no.~4, pp. 357--366, 1980.

\bibitem{hermansky1990perceptual}
H.~Hermansky, ``Perceptual linear predictive (plp) analysis of speech,''
  \emph{The Journal of the Acoustical Society of America}, vol.~87, p. 1738,
  1990.

\bibitem{bottou_global_1997}
L.~Bottou, Y.~Bengio, and Y.~{LeCun}, ``Global training of document processing
  systems using graph transformer networks.'' in \emph{In Proc. of Computer
  Vision and Pattern Recognition}.\hskip 1em plus 0.5em minus 0.4em\relax
  Puerto-Rico., 1997, pp. 490--494.

\bibitem{bottou_stochastic_1991}
L.~Bottou, ``Stochastic gradient learning in neural networks,'' in
  \emph{Proceedings of Neuro-Nîmes 91}.\hskip 1em plus 0.5em minus 0.4em\relax
  Nimes, France: {EC2}, 1991.

\bibitem{lee_speaker-independent_1989}
K.~F. Lee and H.~W. Hon, ``Speaker-independent phone recognition using hidden
  markov models,'' \emph{{IEEE} Transactions on Acoustics, Speech and Signal
  Processing}, vol.~37, no.~11, pp. 1641--1648, 1989.

\bibitem{woodland_large_1994}
P.~Woodland, J.~Odell, V.~Valtchev, and S.~Young, ``Large vocabulary continuous
  speech recognition using htk,'' in \emph{Proc. of ICASSP}, vol.~ii, apr 1994,
  pp. II/125--II/128 vol.2.

\bibitem{morgan_continuous_1995}
N.~Morgan and H.~Bourlard, ``Continuous speech recognition,'' \emph{Signal
  Processing Magazine, {IEEE}}, vol.~12, no.~3, pp. 24 --42, May 1995.

\bibitem{htk}
S.~Young, G.~Evermann, D.~Kershaw, G.~Moore, J.~Odell, D.~Ollason, V.~Valtchev,
  and P.~Woodland, ``The htk book,'' \emph{Cambridge University Engineering
  Department}, vol.~3, 2002.

\bibitem{collobert_torch7_2011}
R.~Collobert, K.~Kavukcuoglu, and C.~Farabet, ``Torch7: A matlab-like
  environment for machine learning,'' in \emph{BigLearn, NIPS Workshop}, 2011.

\end{thebibliography}
\bibliographystyle{IEEEtran}

\end{document}